%% file: main.tex
\documentclass[conference]{IEEEtran}

\input{preamble}

\begin{document}

\title{Aerial Layouting: Design and Control of a Compliant and Actuated End-Effector for Precise In-flight Marking on Ceilings}

\author{\authorblockN{Christian Lanegger\authorrefmark{1},
Marco Ruggia\authorrefmark{1},
Marco Tognon,
Lionel Ott and
Roland Siegwart}
\authorblockA{Autonomous Systems Lab, ETH Zurich, Switzerland}}

\maketitle

\begin{abstract}

Aerial robots have demonstrated impressive feats of precise control, such as dynamic flight through openings or highly complex choreographies. Despite the accuracy needed for these tasks, there are problems that require levels of precision that are challenging to achieve today. One such problem is aerial interaction. Advances in aerial robot design and control have made such contact-based tasks possible and opened up research into challenging real-world tasks, including contact-based inspection. However, while centimetre accuracy is sufficient and achievable for inspection tasks, the positioning accuracy needed for other problems, such as layouting on construction sites or general push-and-slide tasks, is millimetres. To achieve such a high precision, we propose a new aerial system composed of an aerial vehicle equipped with a novel ``smart'' end-effector leveraging a stability-optimized Gough-Stewart mechanism. We present its design process and features incorporating the principles of compliance, multiple contact points, actuation, and self-containment.
In experiments, we verify that the design choices made for our novel end-effector are necessary to obtain the desired positioning precision. Furthermore, we demonstrate that we can reliably mark lines on ceilings with millimetre accuracy without the need for precise modeling or sophisticated control of the aerial robot.

\end{abstract}

\IEEEpeerreviewmaketitle

\input{introduction}
\input{design_objectives}
\input{effector_design}
\input{experiments}
\input{conclusion}

\bibliographystyle{plainnat}
\bibliography{references}

\end{document}

%% file: preamble.tex
\usepackage{times}

\usepackage[numbers]{natbib}
\usepackage{multicol}
\usepackage[bookmarks=true]{hyperref}
\usepackage[nolist,nohyperlinks]{acronym}
\usepackage{graphicx}
\usepackage{siunitx}
\let\labelindent\relax 
\usepackage[inline]{enumitem}
\usepackage{amsmath} 
\usepackage{amssymb} 
\usepackage{mathtools} 

\usepackage{float} 

\usepackage{tabularray}

\usepackage{caption}
\usepackage{subcaption}

\usepackage{adjustbox}
\usepackage{array}
\usepackage{makecell}

\usepackage{cleveref}

\usepackage{pifont}
\newcommand{\cmark}{\ding{51}}%

\hypersetup{
   pdfauthor={Christian Lanegger, Marco Ruggia, Marco Tognon, Lionel Ott, Roland Siegwart},
   pdftitle={Aerial Layouting: Design and Control of a Compliant and Actuated End-Effector for Precise In-flight Marking on Ceilings},
   pdfsubject={Robotics, Flying Vehicle, OMAV, Marking},
   pdfkeywords={Robots; Flying; System; Layouting}
}

\usepackage{xcolor}
\newcolumntype{R}[2]{%
    >{\adjustbox{angle=#1,lap=\width-(#2)}\bgroup}%
    l%
    <{\egroup}%
}

\renewcommand\vec{\mathbf}

\DeclareSIUnit\pixel{px} 

\begin{acronym}
\acro{MAV}{Micro Aerial Vehicle}
\acro{OMAV}{Omnidirectional Micro Aerial Vehicle}
\acro{UAV}{Uncrewed Aerial Vehicle}
\acro{IMU}{Inertial Measurement Unit}
\acro{BCG}{Boston Consulting Group}
\acro{MAE}{Mean Absolute Error}
\acro{MWCE}{Mean Worst Case Error}
\end{acronym}

%% file: introduction.tex
\section{Introduction}

With the increased capabilities and success stories of autonomous systems, more and more applications that previously were out of reach for robotic systems have come within reach. As a result companies invest more heavily in the development of robotic systems. Even the some-times slow to evolve construction field is showing an increased interest in robotics technologies, as a report from the \ac{BCG} \cite{2018bcg} suggests. Some companies have started to develop their own ground robots for specific tasks on constructions sites, for example Hilti who recently presented a semi-autonomous drilling robot\footnote{\href{https://www.hilti.ch/content/hilti/E3/CH/de/products/drilling-and-demolition/semi-autonomer-baustellenroboter-jaibot1.html}{"Semi-autonomous construction robot Jaibot"}, Hilti AG, \textit{last accessed on 25.01.2022}} and Husqvarna who offer remote demolition robots\footnote{\href{https://www.husqvarnacp.com/uk/machines/demolition-robots/}{"Husqvarna DXR remote demolition robots"}, Husqvarna Group, \textit{last accessed on 25.01.2022}}. 

The previously mentioned systems have a limited workspace due to being ground-based robots which prevents them from operating at height or necessitates complicated and heavy structures. Aerial robots, on the other hand, can operate at high altitudes and in areas difficult to reach for ground robots and humans. However, they are especially sensitive to disturbances increasing the challenge to perform very precise and dexterous tasks. Commercially available \ac{MAV} are mainly used for asset management and visual inspection with collision-resilient platforms\footnote{\href{https://flybotix.com/asio-solution/}{"ASIO the revolutionary indoor inspection drone solution"}, Flybotix SA, \textit{last accessed on 25.01.2022}}.
Tasks in construction environments require contact with the environment, a regime which these commercial systems were not designed for. This opens the door for the research of drones capable of safely and precisely interacting with the environment to achieve tasks of interest in the construction sector.

\begin{figure}
    \centering
    \includegraphics[width=1.0\linewidth]{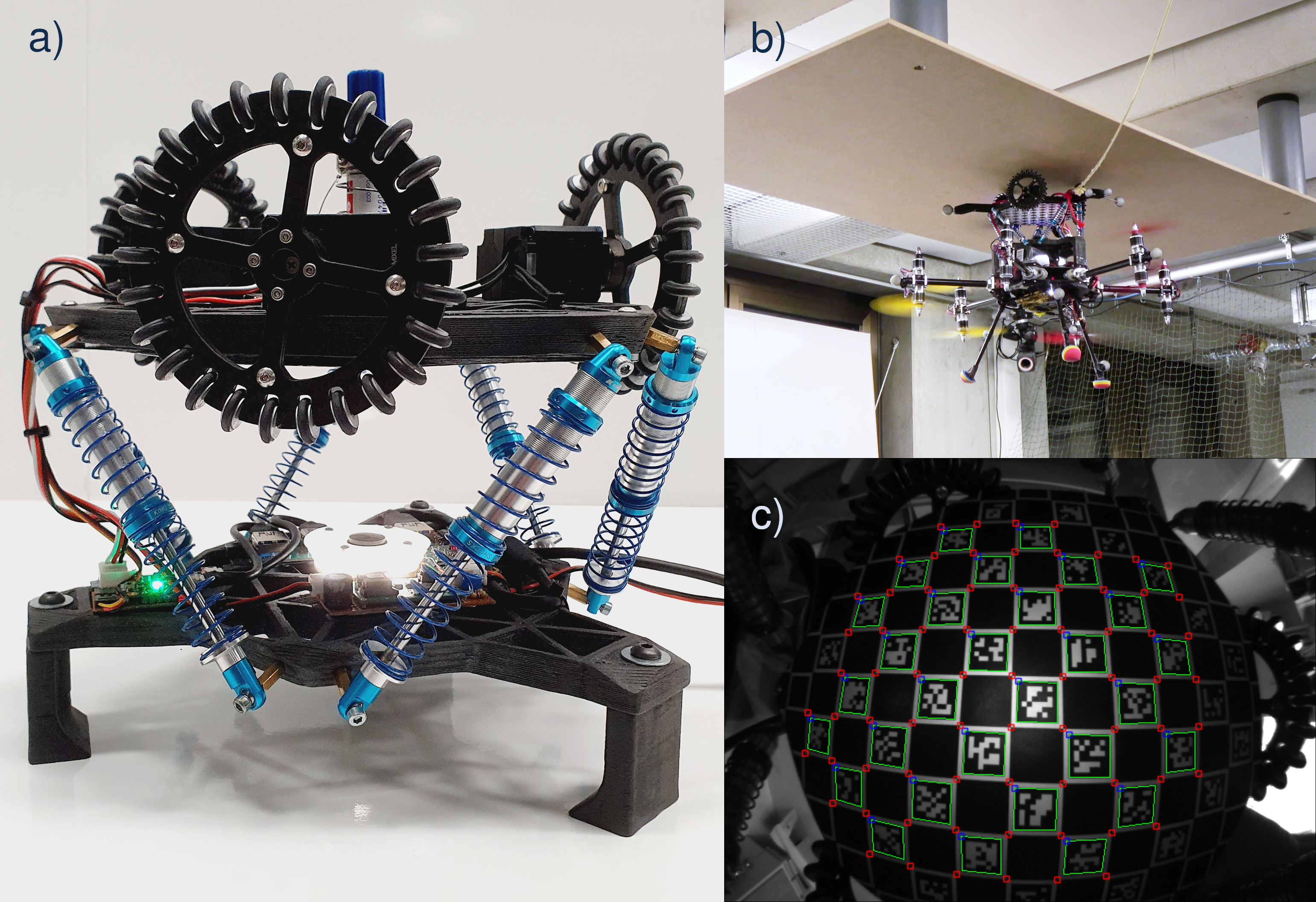}
    \caption{Layouting end-effector based on a passive Gough-Stewart structure (a) for marking on ceilings with aerial vehicles (b). The mobile platform is tracked using an upward facing camera looking at a checkerboard pattern (c). }
    \label{fig:full-system-flying}
\end{figure}

A task of interest in construction is layouting, i.e. the process of marking points, lines, and curves on the ceiling to indicate areas where to drill or anchor components. The manual marking process is repetitive and troublesome, especially at height, and small errors accumulate which can lead to costly remediations, necessitating the markings to be done with millimetre precision. This level of accuracy required from the layouting tool has been rarely achieved with common aerial vehicles. 

In recent years, the research community investigated the usage of aerial manipulators for non-destructive contact-based inspection and manipulation \cite{2021aerialmanip}. To reduce cost and effort involved in bridge inspection several different MAVs for aerial manipulation with a robotic arm attached to the top of the drone \cite{bridgeinspectionocto2017, s20174708} have been proposed. \citet{Jimenez-Cano2015AerialUnderside} also show that by pushing an end-effector against a ceiling the pitch and roll oscillations are reduced, improving the \ac{MAV}'s stability. Other works have shown that aerial manipulators can accurately measure material thickness using ultrasonic probes \cite{voliroTNDE2021} or locate material defects with eddy-current sensors \cite{2019-Tognon_NDT}. Impressive progress has also been made in push-and-slide inspection with aerial systems being able to follow curved surfaces without loosing contact \cite{Trujillo2019NovelIndustry, Nava2020DirectManipulators}. As for most inspection tasks high positional precision is not required the focus of these works mainly lies in the stability of aerial manipulators and the tracking of a desired contact force. These presented systems therefore reach an end-effector accuracy in the order of centimetres.  For layouting tasks, however, centimetre accuracy is insufficient.

To improve precision and disturbance rejection several works propose the use of parallel manipulators. End-effector accuracy can be improved by at least an order of magnitude with the help of a delta-manipulator \cite{deltabodie2021, chermprayong2019integrated}. \citet{Tzoumanikas-RSS-20} showed impressive drawing precision in the millimetre range using nonlinear model predictive control to jointly control \ac{MAV} and delta-arm motion. However, delta-arms add control complexity stemming from the coupled dynamics between aerial vehicle and manipulator. Millimetre accuracy can only be reached if an accurate model of the system is available, making it impractical for complex platforms whose system model has many uncertainties. Additionally, delta arms are only capable of correcting errors in translational motions. Any error in attitude is propagated into the end-effector accuracy. An upward facing manipulator would therefore require the platform to be perfectly level with the ceiling. Marking lines and points with a delta-arm placed on top of a conventional quadrotor would therefore not provide the required accuracy required for layouting.

In this work we present a novel aerial manipulator design consisting of a compliant and actuated end-effector achieving millimetre accuracy (see \Cref{fig:full-system-flying}). In contrast to previous approaches, our design does not rely on the knowledge of an accurate system model nor does it require a sophisticated whole-body control method. We show that the end-effector is able to reach state-of-the-art accuracy without the need for any controller tuning of the aerial vehicle. Similar to humans placing their hand on the surface while writing, the end-effector design exploits multiple contact points to improve stability and precision. Its compliant structure dampens disturbances and also allows the end-effector to be compressed against the ceiling constraining the motion to a plane and helping to stabilize the aerial vehicle. Additional translational actuation at the top of the end-effector corrects for positional inaccuracies and allows for precise surface following. The combination of these features permit accurate end-effector positioning without the need of an extremely precise aerial vehicle.

Our main contribution lies in the design and development of a novel compliant and actuated layouting tool that is based on a Gough-Stewart platform. We set up an optimization framework based on an energy field analysis to find an optimal and stable end-effector geometry. We then implemented a tracking system to precisely estimate the position of the end-effector and introduce a controller to control the actuated portion of the end-effector. Our second contribution lies in the extensive evaluation of the end-effector performance. In initial experiments we  validate our design decisions. Next, we demonstrate repeatibility as well as state-of-the-art accuracy across different trajectories and velocity profiles. These results validate that a smart end-effector design permits millimetre precision without needing a complex control architecture or perfect system model knowledge.

%% file: design_objectives.tex
\section{From Task to Design Objectives}

The goal of this work is to design an aerial layouting system that can be used on construction sites. Its main task is to precisely mark predefined points and lines by sliding a marker along a desired path. The maximum deviation of these markings should be below \SI{5}{\milli\metre}. To further complicate the task lines can be curved, contain corners, and be interrupted.

Instead of designing a complex aerial manipulator requiring sophisticated control and accurate modeling, we believe that a smart mechanical design can provide the desired precision, while simplifying the control system requirements. The result is a rather simple system, whose simplicity increase its robustness, a trait that is desirable in real-world deployments. To achieve our goal, we develop the end-effector around the following four design objectives:
\begin{enumerate}
    \item \textit{Compliance} between the marking element and the aerial platform. This dampens disturbances acting on the aerial vehicle, increasing the precision of the marking;
    \item \textit{Multiple contact points} between ceiling and the end-effector. Maintaining multiple contact points by pushing the end-effector against the ceiling ensures an increased system stability.
    \item \textit{Actuation} to allow controlled movement along the ceiling and further refinement of the end-effector position in order to correct for possible tracking errors;
    \item \textit{Self-containment} of the end-effector preventing the system from relying on a specific aerial vehicle and the need of controller tuning;
\end{enumerate}

%% file: effector_design.tex
\section{End-Effector Design}
With these design objectives in mind we developed an end-effector based on a Gough–Stewart platform \cite{stewart1965platform}. A six-degree of freedom mechanism which consists of two bodies, a usually fixed lower platform and a movable upper platform, both connected by six linear elastic joints as is also shown in \Cref{fig:full-system-flying}(a). This structure provides the necessary compliance to minimize the effects of in-flight disturbances being propagated into the upper platform. 

The upper platform is equipped with three custom-built omni-wheels to provide multiple contact points while still permitting smooth movement in any direction. Each wheel is driven by a servomotor to actively compensate tracking errors and increase precision when in contact with the ceiling. The upper platform is additionally equipped with a retractable permanent marker for marking purposes. An upward facing camera is mounted on the lower end-effector platform to track the relative displacement of the upper platform.  

In the following sections the design is explained in greater detail. We start by introducing the compliant structure in \Cref{sec:stewart-platform} and analyse its stability in \Cref{sec:stability analysis} . Then we present the actuation controller in \Cref{sec:end-effector-control} and discuss the tracking system of the upper platform in \Cref{sec:platform-tracking}. We conclude with \Cref{sec:end-effector-interface} in which we describe the interface between end-effector and aerial vehicle.

\subsection{Gough-Stewart Platform}
\begin{figure}
    \centering
    \includegraphics[width=0.9\linewidth]{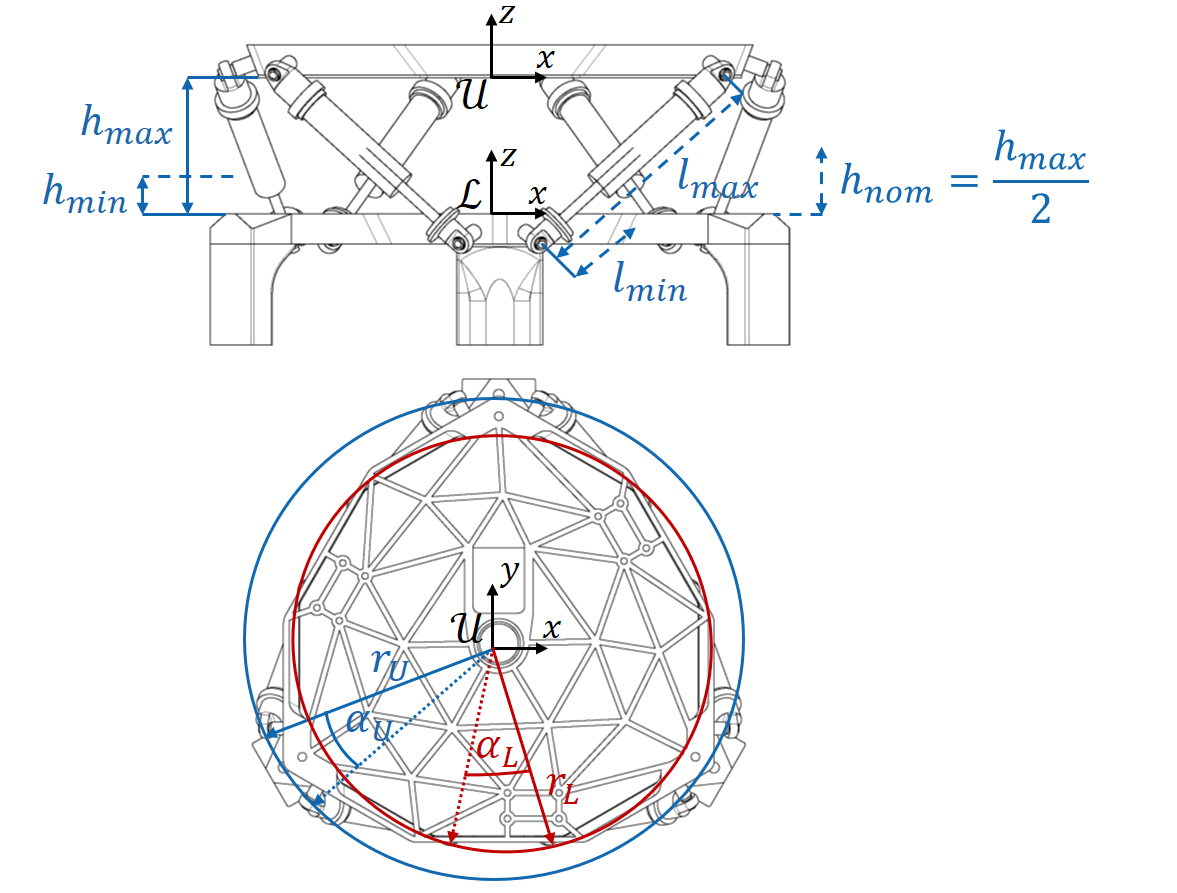}
    \caption{Definition of geometry variables used for the optimization of the Gough-Stewart geometry. $\mathcal{U}$ and $\mathcal{L}$ represent the frames of the upper platform and the lower platform, respectively.}
    \label{fig:geometry_vars}
\end{figure}
\label{sec:stewart-platform}
The Gough-Stewart platform is a parallel manipulator that found its application mainly in flight simulators. As such, the upper platform is actuated using linear actuators. More recently, \citet{hu20186} developed a passive Gough-Stewart structure by replacing the actuated legs with passive springs for vibration isolation. Similarly, our end-effector design uses spring-dampers to connect the two bodies of the Gough-Stewart platform to suppress the vibrations of the aerial vehicle. Additionally, the design allows the mechanism to be compressed which permits the aerial vehicle to push upwards against the ceiling to increase its stability during flight.

The spring-dampers used for the end-effector are off-the-shelf components
\footnote{The exact dampers used are the \textit{RC4WD King Off-Road Dual Spring Shocks} for radio controlled model cars} allowing enough movement of the upper platform to reject in-flight disturbances. Depending on the end-effector geometry the shock absorbers allow for displacement in the range of 2-\qty{4}{\centi\metre}.The spring stiffness was changed such that a force of \qty{15}{\newton} is required to compress the end-effector to its nominal height $h_{nom}$, i.e. center of travel. The required force was chosen as a trade-off between available thrust and ceiling grip. The damping fluid is water as it is empirically found to be the only fluid which, with the given spring stiffness, allowed the upper platform to return to its maximal height.

\subsection{Stability Analysis}
\label{sec:stability analysis}
As the legs of the end-effector are not actuated, the position of the upper platform cannot be actively controlled. The end-effector relies on the spring-dampers pushing the upper platform back to its initial position. Depending on the end-effector geometry this behavior is not always guaranteed. For unstable geometries the upper platform can get stuck at certain locations or even completely drop sideways to the edge of the end-effector's workspace. Such unstable behavior would degrade damping performance and reduce end-effector precision and is therefore not desired.

The end-effector stability is largely dependent on the placement of the spring-dampers. To find a configuration which recovers from most positions and remains stable within the end-effector's workspace, we set up a geometry optimization assigning a scoring metric to different spring-damper placements. Every configuration is evaluated at $h_{nom}$. The upper platform is assumed to be stationary and compressed to $h_{nom}$ by a force along the z-axis for optimization. To assign a stability score to each geometry configuration, the energy required to displace the upper platform in all six degrees-of-freedom, position and rotation, is analysed.  
The energy is calculated as the sum of the energy stored in the spring contraction and the potential energy of the force required to compress the upper platform to $h_{nom}$
\begin{align}
    E\left({}_{\mathcal{L}}\vec{r}_\mathcal{U}, \vec{P}\left(S\right)\right ) = E_{F_z}& \left ( 
    {}_{\mathcal{L}}\vec{r}_\mathcal{U}, \vec{P}\left(S\right)
    \right ) \\ 
    & +
    \sum_{i=1}^{6}{E_{spring}}\left( 
    {}_{\mathcal{L}}\vec{r}_\mathcal{U}, {}_{\mathcal{L}}\vec{p}_l^i, {}_{\mathcal{U}}\vec{p}_u^i \right ) \nonumber
\end{align}
with ${}_{\mathcal{L}}\vec{r}_\mathcal{U} \in \mathbb{R}^{6x1}$ as the pose of the upper platform frame $\mathcal{U}$, given in the lower platform frame $\mathcal{L}$, $S$ as the to be optimized design variables, and $\vec{P} = \begin{bmatrix}{}_{\mathcal{L}}\vec{p}_l^1 & {}_{\mathcal{L}}\vec{p}_u^1 & \hdots & {}_{\mathcal{L}}\vec{p}_l^6 & {}_{\mathcal{L}}\vec{p}_u^6\end{bmatrix} \in\mathbb{R}^{3x12}$ as the stacked vector of all six spring positions at the lower ${}_{\mathcal{L}}\vec{p}_l^i$ and upper platform ${}_{\mathcal{L}}\vec{p}_u^i$ with respect to $\mathcal{L} $. Similar to  \cite{stoughton1993modified}, the design variables are chosen as the following
\begin{equation}
    S = \left\{ r_l, r_u, \alpha_l, \alpha_u, h_{min}, h_{max} \right\}
\end{equation}
where $r_l$ and $r_u$ describe the radii of the lower and upper platform, respectively. The leg positions at the bottom and top body are described by the angles $\alpha_l$ and $\alpha_u$, and the minimal and maximal height of the upper platform by $h_{min}$ and $h_{max}$. The height was chosen as the optimization variable instead of the leg length as the leg length is already fixed by the available hardware. The optimization variables are visualized in \Cref{fig:geometry_vars}.

The stability metric used is the smallest eigenvalue $\lambda_{min}$ of the Hessian of the energy field at the nominal height
\begin{subequations}
    \begin{align}
        H\left (\vec{r}, \vec{P}\left ( S \right ) \right )_{i,j} &= 
        \frac{\partial^2E\left( \vec{r}, \vec{P}\left ( S \right) \right)}{\partial x_i, \partial x_j} \\
        H_{nom}\left(\vec{P}\left(S\right)\right) &=
        \left.
        \begin{matrix}
        H\left(\vec{r}, \vec{P}\left(S\right)\right)
        \end{matrix}\right|_{\vec{r} = \vec{r}_{nom}}
    \end{align}
\end{subequations}
The subscripts have been omitted for readability reasons. The eigenvalues of the Hessian describe the translational and rotational motion of the upper platform. Thus, smaller eigenvalues correspond to a locally flatter energy field, meaning that little energy is needed for displacements along a particular direction. Therefore, the smallest eigenvalue $\lambda_{min}$ corresponds to the worst-case local change along any direction, i.e., least energy needed to displace the mobile platform. The geometry search can now be formulated as a constrained optimization problem:
\begin{equation}
    \begin{aligned}
        \vec{P}^\ast\left(S\right) = 
        \underset{S}{\mathrm{argmax}}\quad & 
        \lambda_{min} \left(H_{nom}\right) \\
        \textrm{s.t.} \quad & \SI{60}{\milli\metre} < r_l, r_u < \qty{130}{\milli\metre} \\
        & \SI{0}{\degree} < \alpha_l, \alpha_u< \SI{30}{\degree} \\
        & h_{min} < \SI{30}{\milli\metre} \\
        & h_{max} > \SI{110}{\milli\metre}  \\
    \end{aligned}
\end{equation}
 
The optimization constraints limit the size of the end-effector and make sure only manufacturable configurations are considered. To find a configuration we finely sample the entire parameter space up to a scale where manufacturing tolerances would negate further optimization and determine $\lambda_{min}$ for every sampled parameter combination. The optimal parameters for the Gough-Stewart platform can be found in \Cref{tab:design-parameters}.
\begin{table}
\centering
\begin{tblr}{lcc}
    \hline
    \textbf{Parameter} & \textbf{Value} & \textbf{Units} \\
    \hline
    $r_l$ & 60 & [mm] \\
    $r_u$ & 115 & [mm]  \\
    $\alpha_l$ & 12.56 & [$^\circ$] \\
    $\alpha_u$ & 12.56 & [$^\circ$] \\ 
    $h$ [min., nom., max.] & [30, 67.1, 94.3] & [mm] \\
    $l_{spring}$ [min., nom., max.] & [80, 100, 120] & [mm] \\
    $k_{spring}$ & 0.2 & [N/m] \\
    \hline
\end{tblr}
\caption{Design parameters of optimized end-effector geometry.}
\label{tab:design-parameters}
\end{table}

It should be noted that the optimization implicitly equates change in position to change in attitude in its score, as the six dimensional energy space is composed of three position and three attitude dimensions. In practice, the translational and rotational motions are scaled such that a \qty{1}{\milli\metre} change in position is equivalent to a \qty{1}{\degree} change in attitude. However, different scalings can put different priorities over position stability vs. attitude stability.

\begin{figure}
    \centering
    \includegraphics[width=1.0\linewidth]{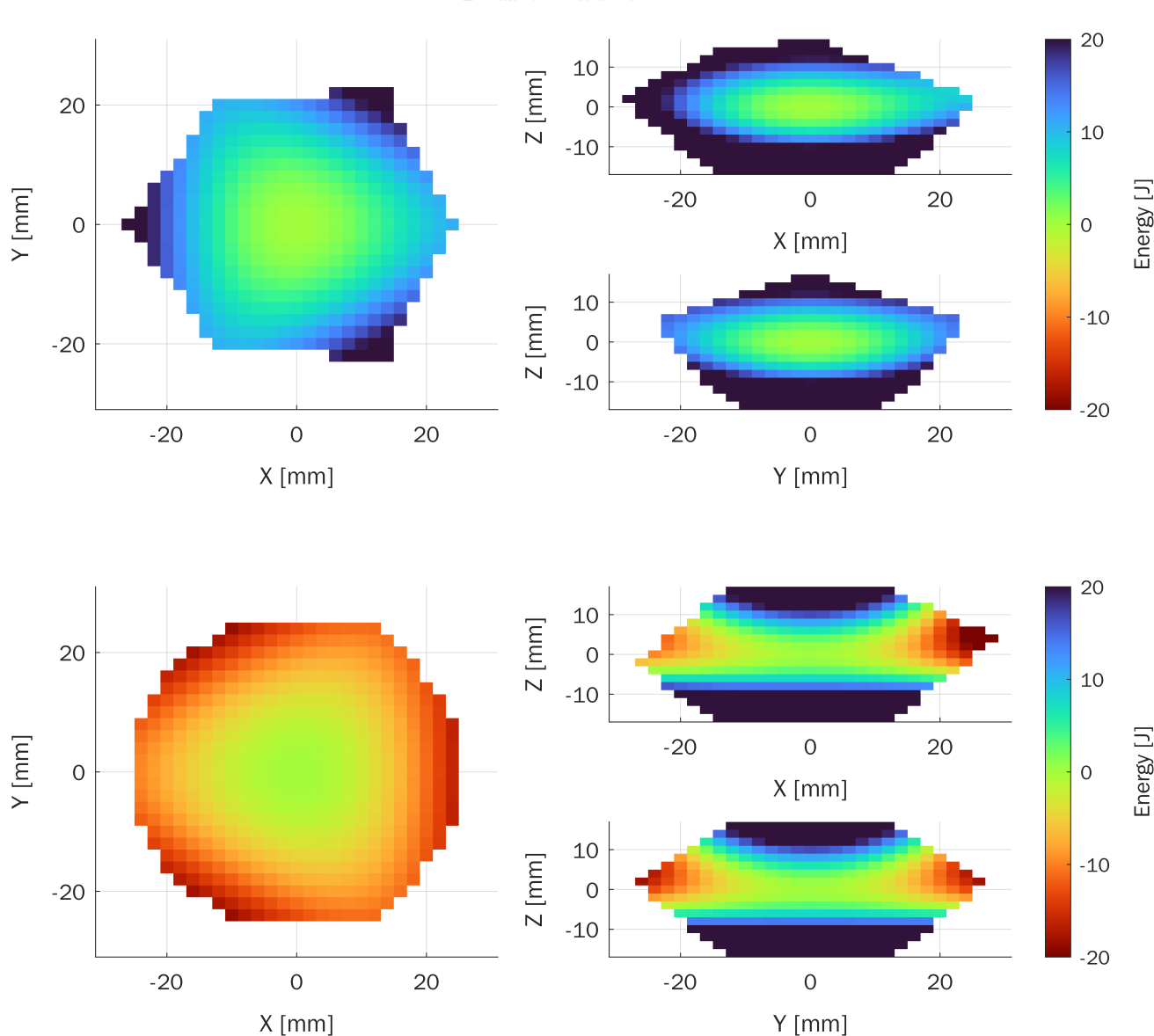}
    \caption{Energy field for the optimized Gough-Stewart structure (top) and for a naive approach (bottom) at the equilibrium position. In contrast to the naive approach, the optimized geometry shows a global minimum at the centered position verifying its stability.}
    \label{fig:heatmaps}
\end{figure}

As the optimization only considers the energy required to displace the end-effector at its nominal height we additionally visually inspected the energy field of the optimized end-effector geometry at different heights. The energy field for the optimal configuration as well as for a naive configuration are shown in \Cref{fig:heatmaps}. The naive configuration represents a Gough-Stewart platform whose bodies are equally sized. The plots show the energy fields at equilibrium height. From the visualization it is clearly visible that the naive structure is only stable when the upper platform is at the center. The energy needed to displace the platform decreases the further the platform moves outwards. This would lead to instability as the upper platform would drop to the edge of its workspace and stay there. For the optimized Gough-Stewart geometry, on the other hand, superior stability behavior can be observed. Here the energy field has a clear global minimum around the center position. The platform is therefore always pushed back to its initial position by the spring-dampers.

\subsection{End-effector Control}
\label{sec:end-effector-control}
The end-effector is equipped with three omni-wheels each mounted to a  servomotor\footnote{The actuators used are \textit{Dynamixel XL430-W250-T}}. This allows precise movement along the ceiling and permits the compensation of positioning errors of the aerial vehicle. 

The omni-wheels are custom-built and made out of carbon fiber to reduce weight. A wheel diameter of \qty{100}{\milli\metre} was chosen providing enough clearance for the marking tool as well as enough positional resolution. The omni-wheels can be seen in \Cref{fig:full-system-flying}(a). 

The upper platform of the end-effector can be modeled as a three-wheeled omnidirectional ground robot. Its configuration is described by its position along the ceiling and its heading with respect to an inertial frame $\mathcal{I}$, both combined in the vector ${}_{\mathcal{I}}\vec{r}_\mathcal{U} = [x \;\, y \,\; \theta]^\top$. To complete its state, we denote the configuration time derivatives with the vector ${}_{\mathcal{I}}\vec{\dot r}_\mathcal{U} = [v_x \,
\; v_y \,\; \omega]^\top$. 
The relation between wheel speeds $[v_1 \,\; v_2 \,\; v_3]^\top$ and upper platform velocity is given by
\begin{equation}
    \begin{bmatrix}
        v_1\\
        v_2\\
        v_3
    \end{bmatrix} = 
    \begin{bmatrix}
          1 & 0 \\
          \cos{\frac{1}{3}\pi} & -sin{\frac{2}{3}\pi} \\
          \cos{\frac{2}{3}\pi} & -sin{\frac{2}{3}\pi} \\
    \end{bmatrix}
    \begin{bmatrix}
    \cos{\left (-\theta\right )} & - \sin{\left (-\theta\right )}\\
     \sin{\left (-\theta\right )} & \cos{\left (-\theta\right )}
    \end{bmatrix}
    \begin{bmatrix} 
    v_x \\
    v_y
    \end{bmatrix}
    \label{eqn:upper_platform_kinematics}
    \end{equation}
with the individual wheel velocities $v_i$ being related to the angular velocity $\omega$ by 
\begin{equation}
 v_i = \omega R \quad , for \quad i = 1,2,3
\end{equation}
with R as the wheel radius. 

To control the upper platform a cascaded controller is used. The inner loop uses the default PI velocity controller with anti-windup of the servo motors. The gains were tuned to be as aggressive as possible without causing instability. The outer loop is designed to be a simple proportional P controller on the position of the upper platform. An integral term is deemed unnecessary as the reference velocity is known and can be passed along as a feed-forward term. As the integrator of the inner loop should be able to prevent static errors in velocity no significant static errors in position should appear.
The controller takes as input the end-effector's desired and current position $x$,$y$ and yaw angle $\theta$ as well as the reference velocity feed-forward term. Based on those measurements, the desired velocity is determined and then allocated to actuator specific velocities $v_1$, $v_2$, $v_3$ using \eqref{eqn:upper_platform_kinematics}. Those are then sent to the build-in velocity controller of each servo motor.

\subsection{Upper Platform Tracking}
\label{sec:platform-tracking}
As the upper platform is not rigidly attached to the aerial vehicle an accurate estimate of the homogeneous transformation $\vec{T}_{\mathcal{LU}}$ from lower $\mathcal{L}$ to upper platform frame $\mathcal{U}$ is required to achieve high precision marking. The estimation accuracy should be within \qty{1}{\milli\metre} so that accurate markings can be made with the end-effector. An external positioning system, like a motion capture system or a total station, would provide such accuracy. During interaction, however, the end-effector is occluded from most directions by the ceiling and the aerial vehicle making it impossible for the positioning system to track the end-effector position accurately. Assuming the pose of the aerial vehicle as well as the transformation $\vec{T}_{\mathcal{BL}}$ between its body frame $\mathcal{B}$ and the lower end-effector platform are known three different approaches to estimate $\vec{T}_{\mathcal{LU}}$ were considered:
\begin{enumerate}
    \item Measuring the extension of each spring-damper using linear potentiometers;
    \item Using a force-torque sensor and stiffness model to measure the forces and torques exerted on the end-effector;
    \item Tracking fiducials on the upper platform using an upward facing camera;
\end{enumerate}
Linear potentiometers would introduce additional friction and more importantly would make the design a lot more complex. Therefore, after a short evaluation this approach was not further pursued.

The use of a force torque sensor seemed also problematic as spring-dampers dissipate some of the force into heat which would lead to inaccurate estimates. Additionally, currently available force-torque sensors of reasonable size do not offer the required accuracy. Assuming that no friction and dampening effects are present and for a spring constant of $k_{spring} = \SI[per-mode = symbol]{0.2}{\newton\per\metre}$, the sensor would require to sense a force of \qty{0.07}{\newton} to measure a translation of \qty{1}{\milli\metre} in either x- or y-direction. This is less than the noise free resolution of force-torque sensors. A Rokubi FT sensor from BotaSystems provides a resolution of \qty{0.3}{\newton}, for instance.
A force-torque sensor would therefore not provide the required accuracy for tracking the position of the upper platform.

Tracking fiducial markers on the end-effector was regarded as the most promising approach. We integrated an \textit{OmniVi-sion OV2311} image sensor packaged in a \textit{See3CAM 20CUG} camera onto the end-effector. The camera was equipped with an \textit{PT-01224XFL} lense. Such a camera setup is in theory able to resolve up to \qty[per-mode = symbol]{0.1}{\milli\metre\per\pixel} at the nominal height $h_{nom}$. On the bottom side of the upper platform a \textit{ChArUco} board was mounted. To equalize the illumination between the two bodies a custom built ring-light was placed around the camera. The \textit{ChArUco} board was tracked using the OpenCV Library \cite{2015opencv}. The tracking system, as well as the camera image of the tracked checkerboard pattern is shown in \Cref{fig:full-system-flying}(a) and \Cref{fig:full-system-flying}(c), respectively. The accuracy of the tracking system was evaluated using a Vicon motion capture system. The results are reported in \Cref{sec:camera-experiments}. 

\subsection{Interface to Aerial Vehicles}
\label{sec:end-effector-interface}
As the end-effector is designed to be a self-contained system, its interface with the aerial vehicle is reduced to a minimum. In fact, the end-effector controller is implemented as an independent unit and as such does not interact with the controller of the aerial vehicle. The end-effector system only requires the current pose estimate of the aerial vehicle and the transformation $T_{\mathcal{BL}}$ for operation. Such a transformation can be obtained beforehand by extrinsically calibrating the end-effector camera to the \ac{IMU} of the aerial vehicle using a toolbox such as Kalibr \cite{kalibr}.

To track a desired trajectory, the same global references in x-, y-direction and yaw are sent to both end-effector and aerial vehicle. The reference Z position of the aerial vehicle is kept constant at the value which compresses the upper platform to its desired height. The Z setpoint is determined using the camera tracking system. During the process of establishing contact and compressing the end-effector the height of the upper platform is tracked. Once the desired height is reached the reference Z position of the aerial vehicle is stored and kept constant throughout the marking process. Currently, no camera feedback is currently used to correct for any error in Z-direction of the aerial vehicle.

%% file: experiments.tex
\section{Experiments}
\subsection{Upper Platform Tracking}
\label{sec:camera-experiments}
\begin{figure}
    \centering
    \includegraphics[width=\linewidth]{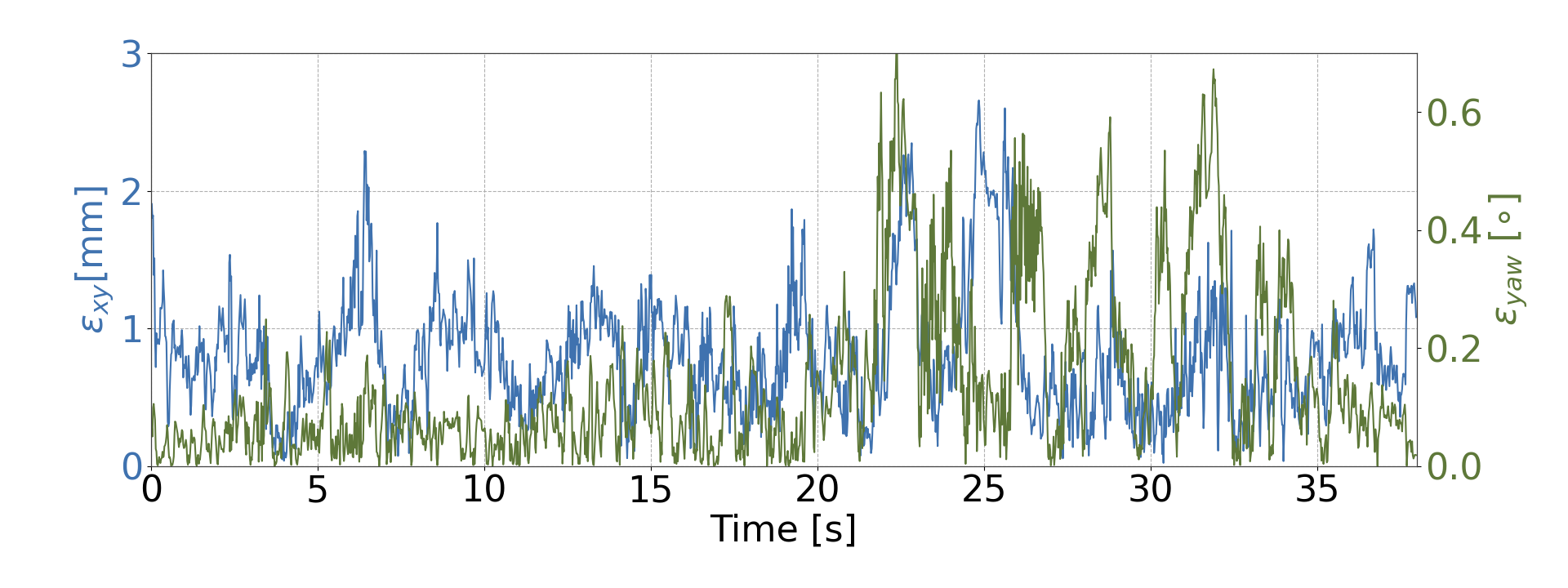}
    \caption{Mean absolute error of the upper platform in the xy-plane (blue) and in yaw (green).}
    \label{fig:tracking-accuracy}
\end{figure}

Prior to evaluating the end-effector performance on an aerial system, we validated that the upper platform tracking provides the required precision to mark locations with millimeters accuracy. To successfully do so we require the position of the platform to be estimated within a millimetre. The accuracy of the tracking system is determined in a static experiment with the help of a Vicon motion capture system. The Vicon system was tracking both the lower and upper body of the end-effector while the upper body was manually moved around. The camera tracking estimates were then compared to the Vicon measurements. The tracking error is shown in \Cref{fig:tracking-accuracy}. On average the camera system is able to estimate the location the upper platform with an error of \qty{0.8}{\milli\metre} in xy-position and \qty{0.2}{\degree} in yaw. The tracking accuracy of the camera system is therefore deemed sufficient for the given application.
\subsection{Experimental Setup}
\label{sec:exp-setup}

The full aerial layouting system has been evaluated with the end-effector mounted on top of a custom build \ac{OMAV}. It is an hexacopter with six additional servomotors allowing to change the tilt angle of every arm individually decoupling the translational movement of the aerial vehicle from its attitude. The \ac{OMAV} carries an \textit{Intel NUC7i7DNBE} running Ubuntu 18.04 as onboard computer, a \textit{Pixhawk4} used for \ac{IMU} measurements and as motor command passthrough, 12 \textit{KDE Direct 2315XF} motors with 9x4.7 propellers, and six \textit{Dynamixel} servos\footnote{The same Dynamixel Model as for the end-effector was used}. The impedance controller developed in \cite{kbodie2019omav}, running on the onboard computer, controls its pose. 

As previously mentioned, the end-effector was designed such that no accurate system model or precise flight controller is needed. To evaluate the end-effector under these conditions, no controller re-tuning was performed for the presented experiments. With the exception of the total mass and center-of-mass offset all control parameters have been kept at the values tuned for flights without payload.

\begin{figure}
    \centering
    \includegraphics[width=\linewidth]{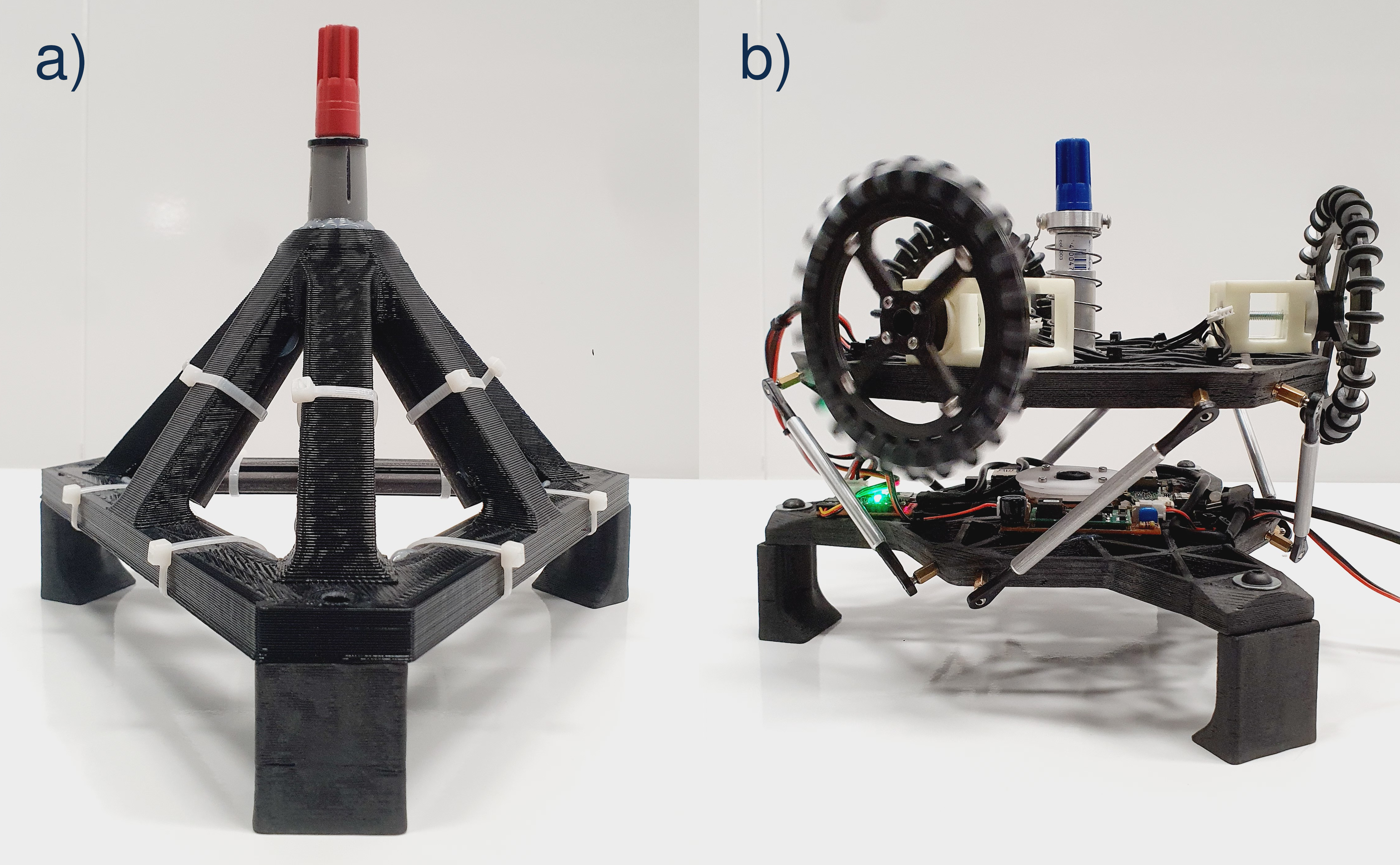}
    \caption{Different end-effector configurations used for design validation experiments. A single-contact end-effector, reverting all design choices (a). On the end-effector friction-less bearings and rigid rods were used to replace servos and spring-dampers, respectively (b).}
    \label{fig:baseline_tools}
\end{figure}
The testbed consisted of an \qty{1.75}{\metre}x\qty{0.9}{\metre}x\qty{0.01}{\metre} MDF-plate fixed parallel to the ceiling and a Vicon Motion Capture system providing external pose estimates for the aerial vehicle. 

Every experiment consisted of the following six steps: 
\begin{enumerate*}[label=(\roman*)]
    \item take off
    \item fly below plate center \label{enum:plate-center}
    \item come into contact with the surface \label{enum:contact-surface}
    \item move to initial position
    \item mark trajectory \label{enum:drawing}
    \item land
\end{enumerate*}.
Step \ref{enum:contact-surface} is assumed completed when the end-effector is compressed to $h_{nom}$. The center of the MDF-plate is determined from Vicon measurements beforehand.

To evaluate the performance of the end-effector we only consider the tracking accuracy during the trajectory marking (step \ref{enum:drawing}). At this stage, the end-effector is located exactly between the wooden plate and the aerial vehicle. It is therefore occluded to the Vicon system from most directions making it impossible to obtain reliable ground-truth measurements. To analyse the end-effector tracking accuracy we therefore used the ground-truth measurements of the aerial vehicle. These were then transformed into the lower platform frame using the position estimate from the upper platform tracking system. The end-effector \ac{MAE} presented in the following experiments does thus not include the tracking error of the camera system. To account for this uncertainty we additionally present a \ac{MWCE} which is defined as the sum of the \ac{MAE} of the end-effector and the camera tracking error (see \Cref{sec:camera-experiments}). Even though, this error does not provide information about the true accuracy of the system it provides an upper bound on the worst case performance.

For the evaluation only x and y end-effector errors, i.e. errors in the plane parallel to the ceiling, were considered. As the end-effector is constrained by the ceiling in height no error in z-direction is present.

The proposed end-effector design is evaluated in three different flight experiments. We start by validating our design decisions in Section \ref{sec:design-eval} through an ablation study in which we remove individual components from the end-effector and evaluate the tracking performance of the reduced system. Then, in Section \ref{sec:tracking-acc} we show repeatability and the overall accuracy by marking a circular trajectory ten times. Additionally, we present results for circular trajectories of different sizes. Finally, in Section \ref{sec:velocity-sweep}, we show consistent tracking performance at different velocities and compare our results to another state-of-the-art approach presented in \cite{Tzoumanikas-RSS-20}.

\subsection{Design Validation}
\label{sec:design-eval}
\begin{table*}
\centering
\begin{adjustbox}{max width=\textwidth}
\begin{tblr}{ |rc|c|c|c|c|c|c|c|c|c|c|c| }
 \hline
  & &
  &  &  &  &  &
 \SetCell[c=3]{c} End-Effector & & &
 \SetCell[c=3]{c} OMAV \\
  & &
  &  &  &  &  &
 \SetCell[c=3]{c} xy-error [mm] & & &
 \SetCell[c=3]{c} xyz-error [mm] \\
 Experiment & \# &
 Contact & Wheels & Compliance & Actuation & Feedback &
 MAE (MWCE) & STD & P\textsubscript{90\%} &
 MAE & STD & P\textsubscript{90\%} \\
 \hline
 Free-Flight &
 \ding{172}          &        &        &        &        & 
                     & 22.6 (-)  & 13.0   & 43.5   & 21.  6 &  9.0 & 33.8 \\
 Single Contact &
 \ding{173}          & \cmark &        &        &        &  
                     & 40.5 (-)   & 27.5   & 87.1   & 21.0   &  7.7 & 31.4 \\
 \hline[dashed]
 Friction-less &
 \ding{174}          & \cmark & \cmark &        &        &  
                     & 14.5 (15.3)  &  7.9   & 25.0   & 13.2   &  5.8 & 20.8 \\
 Spring-Dampers &
 \ding{175}          & \cmark & \cmark & \cmark &        & 
                     & 33.1 (33.9)  & 15.3   & 54.5   & 15.4   &  6.2 & 23.6 \\
 Feedforward Control &
 \ding{176}          &  \cmark & \cmark & \cmark & \cmark &  
                     & 33.2 (34.0)   & 25.1   & 69.8   & 22.4   & 15.8 & 44.5 \\
 \hline[dashed]
 Actuation &
 \ding{177}          & \cmark & \cmark &        & \cmark & \cmark  
                     &  1.3 (2.1)  &  0.8   &  2.4   &  \textbf{7.7} &  \textbf{1.5} &  \textbf{9.8} \\
 Full System &
 \ding{178}          & \cmark & \cmark & \cmark & \cmark & \cmark  
                     &  \textbf{1.0 (1.8)}   &  \textbf{0.5}   &  \textbf{1.7} & 12.2 &  3.4 & 15.6 \\
 \hline
\end{tblr}
\end{adjustbox}
\caption{\ac{MAE}, standard deviation, and the 90th percentile for the end-effector and aerial vehicle for every individual design validation experiment.}
\label{tab:design-validation-results}
\end{table*}

To understand the impact of every individual design choice we conducted experiments of the same circular trajectory for different end-effector setups. The circle had a radius of \qty{250}{\milli\metre} and was tracked using a maximal velocity of \qty[per-mode = symbol]{5}{\centi\metre\per\second} and maximal acceleration of \qty[per-mode = symbol]{2.5}{\centi\metre\per\second\squared}. The components used in every individual experiment as well as the corresponding tracking errors are presented in \Cref{tab:design-validation-results}.

Two experiments serve as baseline. They track the circular trajectory
\begin{enumerate}
\item in free flight showing the actual flight performance of the \ac{OMAV} with the end-effector mounted on top. On average a precision of \qty{22.6}{\milli\metre} was achieved with \qty{90}{\percent} of the measured error lying within \qty{43.5}{\milli\metre}. This large error stems mainly from two sources. First being the model uncertainties of the rather complex aerial vehicle (backlash in tilt-angle servos, actual motor thrust, etc.), and second being the badly tuned controller parameters for the used setup. 
\item using a single point of contact end-effector with the contact point being the marker itself . It does not have any compliance with respect to the \ac{MAV} and does not have any actuators, nor associated feedback (see \Cref{fig:baseline_tools}(a)). As such it undoes all concept decisions. To reduce friction, the upward thrust was reduced to \qty{2}{\newton} which is approximately the minimum force required to maintain contact to the ceiling. Nevertheless, the  static friction between marker and ceiling was still enough to cause jerky movements and constant overshooting of the tool tip resulting in a worse performance compared to the first experiment, which we recall does not include contact. 
\end{enumerate}

Rigid rods and dummy servos with fricton-less bearings have been manufactured to allow us to individually revert the compliance and actuation of the end-effector. Different combinations of these two components were used to evaluate the performance of the same end-effector with less-functionalities. In experiment \ding{174} both components were used adding three friction-less contact points between ceiling and end-effector (see  \Cref{fig:baseline_tools}(b)). This alone almost halves the \ac{MAE} of both end-effector and \ac{OMAV} compared to free flight proving that additional contact points can be exploited to stabilize the \ac{OMAV}.
However, adding compliance to such a system, will not improve tracking precision of the end-effector, even though the accuracy of the \ac{OMAV} stays the same. Without actuation the stiffness of the compliant structure in the XY-plane is not large enough to keep the upper platform of the end-effector at its center position. It gets therefore dragged behind by the \ac{OMAV} while moving along the ceiling. Equally bad performs a compliant end-effector with an actuated upper platform that has no notion about the upper platform's position. Without feedback the end-effector controller controls the omni-wheels only in feedforward fashion. The end-effector is therefore unable to correct for drifting position leading to a bad tracking performance as \ding{176} shows.

The main performance gain can be obtained by using an end-effector
with feedback controlled actuators. Experiments \ding{177} and \ding{178} suggest that such an end-effector reaches an accuracy of less than \qty{1.5}{\milli\metre} improving the accuracy to any previous experiment by an order of magnitude. The results suggest compliance is not strictly required. A main performance increase is already reached by replacing friction-less contact-points (Experiment \ding{174}) with precisely controlled wheels (Experiment \ding{177}). The main reason behind this is that the actuators help to increase friction in unwanted directions while still keeping the friction low in the direction of the trajectory. Therefore, the better the end-effector's actuators can track the reference trajectory the more accurate the system becomes. Nevertheless, adding compliance further reduces the error in our experiments. We believe that under laboratory conditions a system without compliance and a better performing flight controller achieves the same tracking performance as our end-effector design. However, a compliant end-effector would still provide the most accurate tracking in the presence of unforeseen disturbances, such as wind gusts.

\subsection{Tracking Accuracy \& Repeatability}

\label{sec:tracking-acc}
\begin{figure*}
    \centering
    \includegraphics[width=\textwidth]{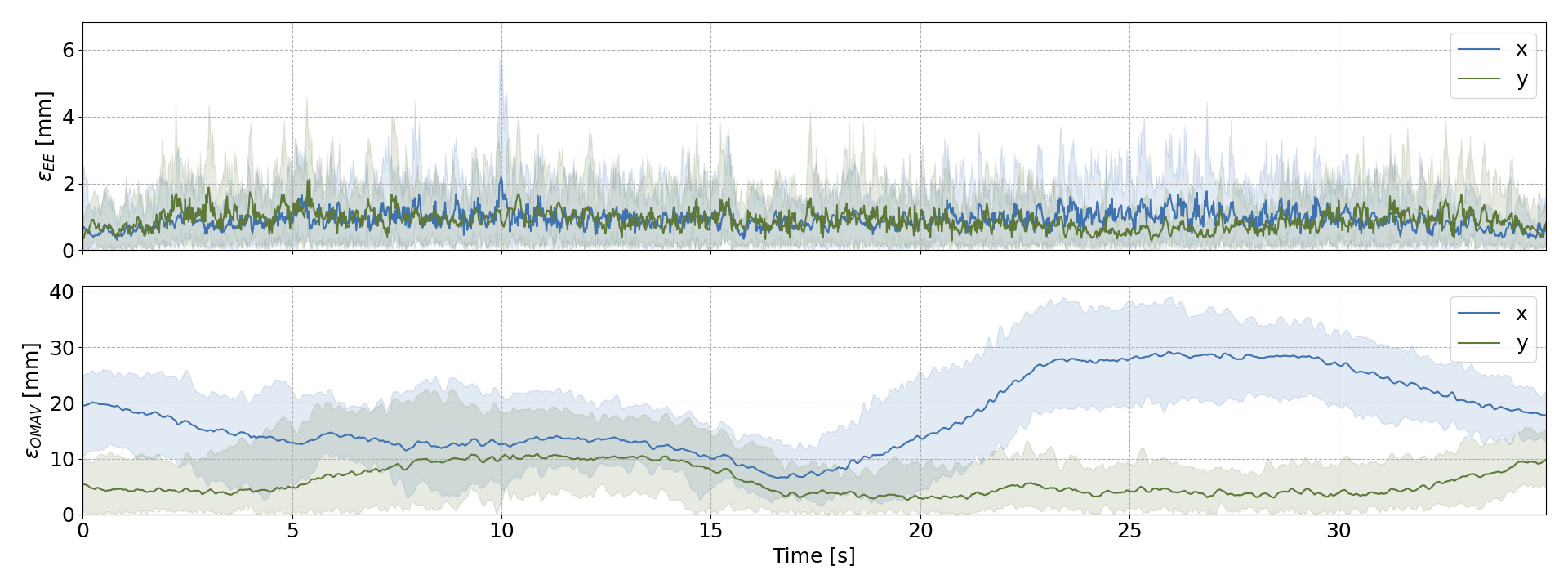}
    \caption{Average tracking error of the end-effector (top) and \ac{OMAV} (bottom) for ten marking sequences of a circle with a radius of \qty{250}{\milli\metre}. The colored regions represent the bound between the 10th and 90th percentile of the tracking error.}
    \label{fig:circle-mean-std}
\end{figure*}
As suggested by \citet{suarez2020benchmarks}, we evaluate our trajectory accuracy by marking the same circular trajectory ten times. Again, a maximum reference velocity of \qty[per-mode = symbol]{5}{\centi\metre\per\second} and maximum acceleration of \qty[per-mode = symbol]{2.5}{\centi\metre\per\second\squared} was used. \Cref{fig:circle-mean-std} shows the mean error and the bound between the 10th and 90th percentile error for both, the end-effector and the aerial vehicle. 
The \ac{MAE} error of the end-effector rarely exceeds \qty{2}{\milli\metre} in either direction, while the \ac{MAE} of the \ac{OMAV} ranges between a few millimetres and \qty{3.5}{\centi\metre}. It can be seen that the end-effector error remains almost constant for the whole trajectory. Even for sections in which the \ac{OMAV} tracking performance degrades (between $t=20$ and $t=30$) no reduced end-effector accuracy can be observed. This shows that the end-effector successfully manages to repeatedly correct for the inaccuracies of the aerial vehicle.

\begin{figure}
    \centering
    \includegraphics[width=0.5\textwidth]{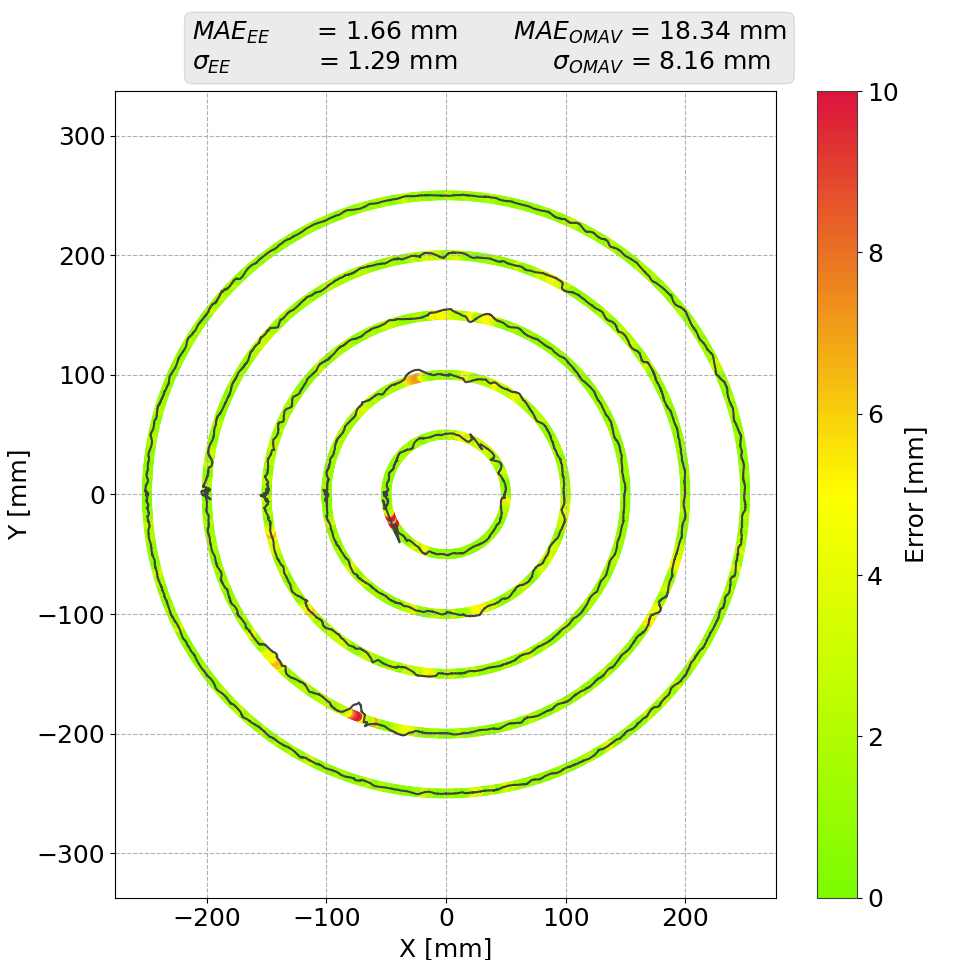}
    \caption{Absolute tracking error of end-effector for different radii. Small segments with errors exceeding \qty{10}{\milli\metre} come from jumps in the external pose estimation resulting in jerky movements of the flying vehicle.}
    \label{fig:circle-size-sweep}
\end{figure}

Additionally, we show that similar tracking accuracies can be reached for trajectories of different sizes. In \Cref{fig:circle-size-sweep} we show the absolute tracking error for circles with radii $r=\left\{50,100,150,200,250\right\}\SI{}{\milli\metre}$. The results are consistent with the previous experiments proving that also tighter curves can be marked without loss of accuracy. The small sections in which the absolute error approaches \qty{10}{\milli\metre} are a result of bad pose estimates fed into the onboard state estimation of the flying vehicle. This lead to jerky flight maneuvers which the end-effector was not able to fully counteract as the upper platform reached its maximal displacement.

\subsection{Velocity Sweep}
\label{sec:velocity-sweep}
\begin{figure*}
    \centering
    \includegraphics[width=\linewidth]{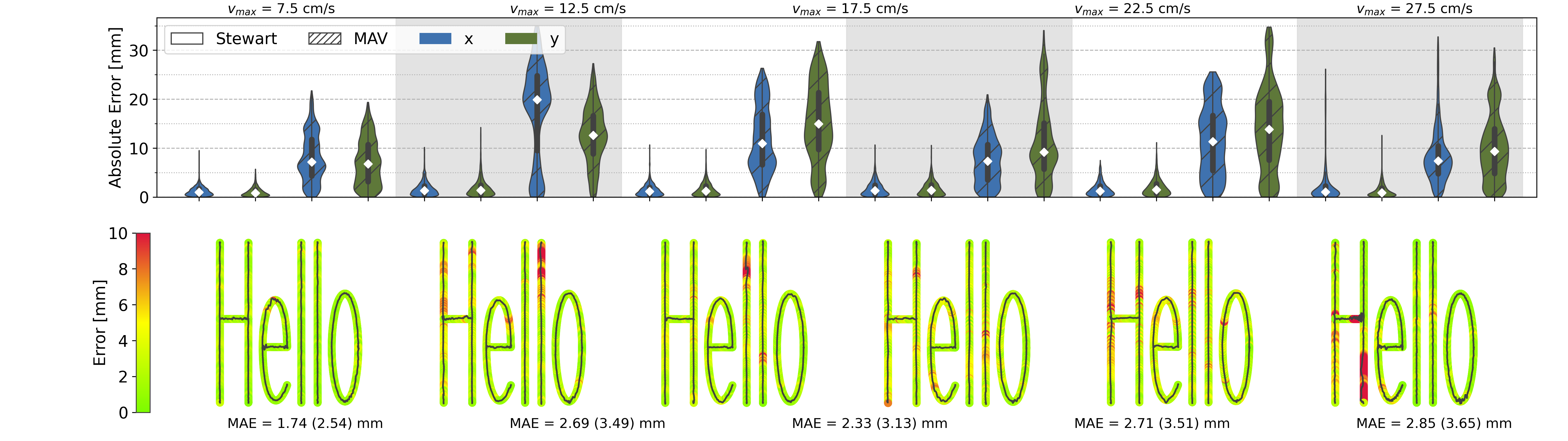}
    \caption{Violin plots (top) of the absolute tracking error of the end-effector and \ac{OMAV} with corresponding visual error (bottom) for 5 \texttt{hello} trajectories. The maximal velocity and acceleration was increased with every iteration (left to right).}
    \label{fig:hello-plots}
\end{figure*}

To evaluate the impact of different end-effector velocities on the tracking performance we reproduced the velocity sweep experiments from \citet{Tzoumanikas-RSS-20} using our system. We tracked a \texttt{Hello} trajectory with similar velocity and acceleration profiles, namely $v_{max} = \left\{7.5, 12.5, 17.5, 22.5, 27.5 \right\}\SI[per-mode = symbol]{}{\centi\metre\per\second}$ and $a_{max} = \left\{3.75, 6.25, 8.75, 11.25, 13.75 \right\}\SI[per-mode = symbol]{}{\centi\metre\per\second\squared}$. The violin plots for the end-effector and aerial vehicle tracking accuracy are shown in \Cref{fig:hello-plots}. It can be seen that the end-effector successfully reduces the average error of the \ac{OMAV} by an order of magnitude for any velocity. By observing the error distributions it becomes clear that the end-effector efficiently rejects all external disturbances. The error of the \ac{OMAV} is rather uniformly distributed between \qty{0}{\milli\metre} and \qty{35}{\milli\metre}, the density of the end-effector error distribution is the densest below \qty{5}{\milli\metre}. Only very few measurements led to a larger error. These larger errors are again caused by jumps in the external pose estimation causing jerky motion of the flying vehicle. The measurements provided by the Vicon system were often unreliable at the most leftward vertical line and the upper part of both \texttt{l} segments.

We achieve similar end-effector tracking accuracies as presented in \cite{Tzoumanikas-RSS-20} who report for the \texttt{hello} trajectory an error between \qty{2.1}{\milli\metre} and \qty{2.8}{\milli\metre}. In our case, the \ac{MAE} stays below \qty{3}{\milli\metre} and below \qty{4}{\milli\metre} for the \ac{MWCE} for all velocity profiles with the lowest speed reaching an \ac{MAE} of \qty{1.74}{\milli\metre}. This proves that our system is able to consistently track even complex trajectories with a similar precision as the state-of-the art without the need for an accurate system model or complex control architecture.

%% file: conclusion.tex
\section{Conclusion} 
\label{sec:conclusion}

This work presented the design and evaluation of a novel end-effector designed for high-precision ceiling-based interaction tasks, focusing on marking points and lines for layouting on construction sites. The system achieves high accuracy through the use of multiple contact points, compliance, and end-effector actuation. The benefit of incorporating these features into the design is that neither precise modeling nor sophisticated control of the aerial vehicle is needed, which can be challenging to achieve for complex systems.

In ablation studies, we verified the hypothesis that multiple contact points, that allow the aerial vehicle to push against a rigid body and stabilize itself, significantly improve end-effector positioning accuracy. Furthermore, we showed that closed-loop control of the actuated end-effector provides additional accuracy benefits when executing complex trajectories while in contact. We further evaluated our proposed design through a series of experiments achieving millimetre accuracy on trajectories of varying complexity and end-effector velocities.

In the future, we plan to expand the workspace of the end-effector to inclined and curved surfaces and test the end-effector with different inspection tools and aerial vehicles. Finally, we will investigate if a whole-body control approach provides additional benefits to the complete system.